\documentclass[10pt,twocolumn,letterpaper]{article}
\pdfoutput=1
\usepackage{cvpr}
\usepackage{times}
\usepackage{epsfig}
\usepackage{graphicx}
\usepackage{amsmath}
\usepackage{amssymb}
\usepackage{booktabs}
\usepackage{subcaption} 
\usepackage{adjustbox}
\usepackage{hyperref}
\usepackage{tikz}
\usetikzlibrary{shapes,arrows}
\usepackage{amsmath,bm,times}
\usetikzlibrary{positioning}

\cvprfinalcopy % *** Uncomment this line for the final submission

\def\cvprPaperID{2314} % *** Enter the CVPR Paper ID here

\ifcvprfinal\fi
\begin{document}

%%%%%%%%% TITLE
\title{PN-Net:  Conjoined Triple Deep Network for \\Learning Local Image Descriptors}

\author{Vassileios Balntas\\
University of Surrey\\
{\tt\small v.balntas@surrey.ac.uk}
\and
Edward Johns\\
Imperial College London\\
{\tt\small  e.johns@imperial.ac.uk}
\and
Lilian Tang\\
University of Surrey\\
{\tt\small h.tang@surrey.ac.uk}
\and
Krystian Mikolajczyk \\
Imperial College London\\
{\tt\small k.mikolajczyk@imperial.ac.uk}
}

\maketitle
%\thispagestyle{empty}

%%%%%%%%% ABSTRACT
\begin{abstract}
In this paper we propose a new approach for learning local descriptors for matching  image patches. 
It has recently been demonstrated that descriptors based on
convolutional neural networks (CNN) can significantly improve the matching performance. Unfortunately  their computational complexity is prohibitive for any practical application. We address this problem and
propose a CNN based descriptor with improved matching performance, significantly reduced training and execution time, as well as low dimensionality. We propose to train the network with triplets of patches that include a positive and negative pairs. To that end we introduce a new loss function that exploits  the relations within the triplets.
We compare our approach to recently introduced MatchNet and
DeepCompare and demonstrate the advantages of our descriptor in terms
of performance, memory footprint and speed i.e. when run in GPU, the
 extraction time of our 128 dimensional feature is comparable to the fastest available
binary descriptors such as BRIEF and ORB. 
\end{abstract}

%%%%%%%%% BODY TEXT
\section{Introduction}

Finding correspondences between images via local descriptors is one of the most extensively studied problems in computer vision due to the wide range of  applications. The field has witnessed several breakthroughs in this area such as SIFT~\cite{Lowe:2004:DIF:993451.996342}, invariant region detectors~\cite{mikolajczykIJCV2004}, fast binary descriptors~\cite{Calonder:2010:BBR:1888089.1888148}, optimised descriptor parameters \cite{WHB09,simonyan2014learning}  which have made a significant and wide impact in various computer vision tasks.  
Recently end-to-end learnt descriptors~\cite{FDB14,simo2015deepdesc,ZagoruykoCVPR2015,Han_2015_CVPR} based on CNN architectures were demonstrated to significantly outperform state of the art features. This was a natural adoption of CNN  to local descriptors as deep learning had already been shown to significantly improve in many computer vision areas \cite{lecun2015deep}. 
However, the performance improvements with CNN based descriptors come at the cost  in terms of of extensive training time, computation,  size of the annotated data as well as significantly larger dimensionality of the feature vector. For example, days of training with GPU on 100s of thousands of training patches  are reported in \cite{FDB14,simo2015deepdesc,ZagoruykoCVPR2015,Han_2015_CVPR}, and descriptor dimensionality reaches up to 4096. Moreover, slow execution time i.e. descriptor extraction, even using GPUs, negatively outweighs the benefits of improved matching.  Even though the efficiency and dimensionality is improving with new methods, it is still far off descriptors such as BRIEF i.e. 8 bytes, 3$\mu s$ per descriptor. 

Another issue in the area of matching patches is the limited benchmark data. Typically used Oxford data~\cite{schmid2003performance} was designed a decade ago and can be considered very small for today's standards. Also, its original evaluation protocol included detection of interest points which currently are less often used in the evaluations. Hence, different protocols and evaluation measures are adopted in various papers which make a comparative study inaccurate. Patch data from~\cite{BHW10} with well defined groundtruth is more convenient to use but the reported error has decreased  significantly over past years such that the margin for improvement is very small. Furthermore the training and testing is done on the data from similar distribution and over-fitting may occur.

In this paper we propose a CNN based descriptor that improves the performance of recent methods, reduces  matching error from $26\%$ (SIFT) to $\approx 7\%$, it is of the same dimensionality as SIFT, its extraction time is 40 times faster than SIFT and only 3 times slower than BRIEF, and single epoch training time is 2$min$. We propose to train the network with {\em Positive} and {\em Negative} pairs formed by triplets of patches, hence our network is termed PN-Net. We introduce a new loss function, which we call SoftPN, to simultaneously exploit the constraints given by the positive and negative pairs. We compare our network and the loss function to other CNN based approaches and demonstrate the improvements. We extend the Oxford data with new image sequences and modify the evaluation protocol such that the data can be used in a similar way to the one from~\cite{WHB09}, yet it preserves the advantages of having the various type of noise separated for more detailed analysis of descriptors. Together with the patch data it allows to test the generalisation properties of the evaluated methods. We perform extensive evaluation and comparison to the state of the art descriptors and demonstrate the improvements in matching performance, extraction efficiency, dimensionality, and training time.

\section{Related work}

%\subsection{The beginning: manually designed features}

The design and implementation of local descriptors has undergone a remarkable evolution over the past two decades ranging from differential or moment invariants, correlations, histograms of gradients or other measurements, PCA projected patches etc. An overview of  pre-2005 descriptors with SIFT \cite{Lowe:2004:DIF:993451.996342} identified as the top performer can be found in \cite{schmid2003performance}. Its benchmark data accelerated the progress in this field and there have been a number of notable contributions, including recent DSP-SIFT 
\cite{DBLP:journals/corr/DongS14}, falling into the same category of descriptors as SIFT but the improvements were not sufficient to replace SIFT in various applications. The research focus shifted to improve the speed and memory footprint e.g. as in BRIEF \cite{Calonder:2010:BBR:1888089.1888148} and the follow up efforts.  Introduction of datasets with correspondence ground truth \cite{WHB09} stimulated development of learning based descriptors which try to optimise descriptor parameters and learn projections or distance metrics \cite{BHW10,simonyan2014learning} for better matching. 

End-to-end learning of patch descriptors using CNN has been attempted in several works \cite{FDB14,ZagoruykoCVPR2015,simo2015deepdesc,Han_2015_CVPR} and consistent improvements were reported over the state of the art descriptors.  It was shown in \cite{FDB14} that the
features from the last layer of a convolutional deep network trained
on ImageNet \cite{ILSVRC15} can outperform  SIFT. Furthermore,  training a siamese deep network with hinge loss in \cite{ZagoruykoCVPR2015,simo2015deepdesc,Han_2015_CVPR} (two CNN's sharing the same
weights) based on positive and negative
patch pairs, resulted in significant improvements in matching performance. Explicit metric learning is often performed in such descriptors to classify similar and dissimilar pairs. This may lead to sub-optimal performance if such learnt representation is used as a descriptor for a different task.  
 
It is well known that careful selection of training data may lead to significant performance increase.
Inspired by relevant methods in SVM based classifiers \cite{lsvm-pami}, the approach from \cite{simo2015deepdesc} proposes to improve learning by  mining  and re-training the network with hard training examples.
Many similar ideas can be found in the area of distance metric learning \cite{Bellet2014} which also exploits various methods of sampling data points to train better projections. In particular Linear Discriminant Embedding, Marginal Fisher Analysis, Neighbourhood Component Analysis or Large Margin Nearest Neighbour focus on exploiting nearest neighbours in training examples and their relations rather than treating all data points equally. A notable example in the context of local descriptors is the  nearest neighbour ratio \cite{Lowe:2004:DIF:993451.996342}  used for matching instead of the absolute Euclidean distance. Similarly, bootstrapping techniques often rely on identifying false positives and false negatives and improve classifiers by re-training on those.
The idea of guiding the learning process simultaneously by positive
 and negative constraints was successful exploited in PN-Learning \cite{kalalCVPR2010} in patch based online learnt object detector. 
 
We build on the top of these results and propose PN-Net that exploits
the positive and negative relations within triplets of training
examples in contrast to pairs in siamese networks. A similar idea was
recently investigated in the context of image categorisation into 10
object classes, but the reported improvements were marginal~\cite{DBLP:journals/corr/HofferA14}. Our triplet network structure and
the loss function is different and a patch can be considered as an
independent object class, thus the typical matching problem  includes 1000s of such classes. We also
design a new loss function termed {\em SoftPN}, which is inspired by SoftMax ratio and hard negative mining from~\cite{simo2015deepdesc}. As we demonstrate in the experiments our method leads to significant improvements in terms of matching performance, dimensionality,  and both training and test speed.

\section{PN-Net}
In this section, we present our feature descriptor learning method,
and we give a brief analysis of its strengths against the previously
used CNN architectures and loss functions.

\subsection{Overview of the network architecture}
Our goal is to compute a representation vector $D(p) \in \mathbb{R}^{\mathcal{D}}$ of image  
 patch $p \in \mathbb{R}^{N \times N}$. Descriptor vector $D(p)$ results from the final layer
of a convolutional neural network where the layer size matches the feature dimensionality. 
% Unlike \cite{Han_2015_CVPR} or \cite{ZagoruykoCVPR2015} we are not interested in producing a final metric layer that learns to output patch pair distances for matching purposes. 
In contrast to \cite{Han_2015_CVPR} or \cite{ZagoruykoCVPR2015}, which include metric learning, our goal is to generate descriptors that can be used in traditional matching setup i.e. with the $L_2$ distance. This has the advantage of opening the application range to
various well-studied techniques such as KD-Trees or approximate nearest neighbour search. %\cite{simo2015deepdesc}.

Previous work on deep learning of feature descriptors has
been based on the {\em siamese networks} as illustrated in Figure \ref{fig:architectures}(top). Such networks
consist of two CNNs which accept two parallel inputs and share parameters across networks. The loss function is optimised based on
the output of the two networks according to their distinct inputs. A single distance between a pair of patches is considered for every training example. This architecture is used to extract descriptors showing state-of-the art matching performance in \cite{simo2015deepdesc,ZagoruykoCVPR2015}. 

\begin{figure}[t]
    \centering
    \begin{subfigure}[b]{1.0\columnwidth}
        \centering
        \includegraphics[width=\columnwidth]{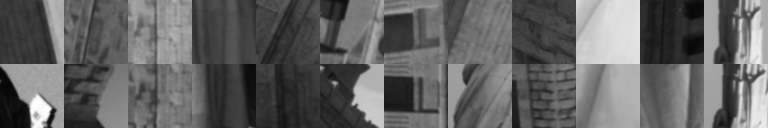}\\
         \vspace{3pt}
        \includegraphics[width=\columnwidth]{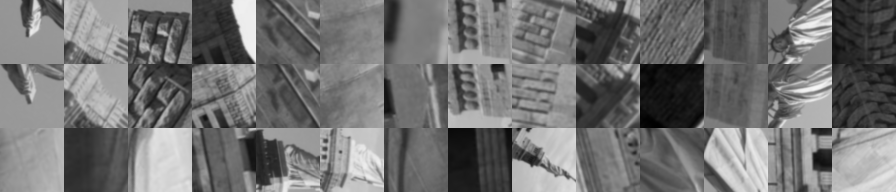}
    \end{subfigure}
    \caption{ Siamese networks are trained with positive and
      negative pairs of patches (top). 
       Our network is trained with
      triplets of patches (bottom), two extracted from the same point representing a positive match example, and
      one from a different point in space giving  two negative
      match examples per triplet.}
\label{fig:pairs_and_triplets}
\end{figure}

 Architectures exploiting three parallel inputs have recently been investigated in \cite{DBLP:journals/corr/WangSLRWPCW14} and
\cite{DBLP:journals/corr/HofferA14} in the context of ranking and classification. Their loss function makes use of a triplet of images
 where  two images are from the same class, and one is from a different class.  Inspired by these techniques we propose an approach to learning local feature descriptors for matching patches. We make use of a triplet of patches, where two of them are positive patches from two views of the same point in the 3D space, and the third one is a negative patch extracted from a
different point in space. The loss function is then based on three distances considered simultaneously for every training example formed from the triplet. The proposed network is shown in Figure \ref{fig:architectures}(bottom).
 Examples of the positive and negative training pairs used in previous works as well triplets exploited in our network  are shown in Figure \ref{fig:pairs_and_triplets}.

%In order to generate our dataset, we utilize the Photo-Tourism dataset \cite{BHW10} which consists of a sequence of 3D points together with several patches from different views of each point. In order to generate triplets we randomly sample two patches form the same 3d point, and we append a third patch from a different 3D point. 

\tikzstyle{cnn} = [draw, fill=blue!20, rectangle, 
    minimum height=3em, minimum width=3em]
\tikzstyle{loss} = [draw, fill=green!20, rectangle, 
    minimum height=5em, minimum width=3em]

\begin{figure}[t]
\centering

% siamese
\begin{tikzpicture}[auto, node distance=2cm,>=latex']
  \node (patch1) at (0,0) {\includegraphics[width=.05\textwidth]{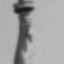}};
  \node (patch2) [below = 0.45cm of patch1] {\includegraphics[width=.05\textwidth]{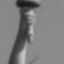}};

  \node (cnn1) [cnn, right = 0.5cm of patch1] {CNN};
  \node (cnn2) [cnn, below =0.5cm of cnn1] {CNN};

  \node (loss) [loss, below right=-0.5cm and 1.2cm of cnn1] {\begin{tabular}{c}
                                                               $Loss:
                                                               \;
                                                               function
                                                               \;of$\\ 
                                                               $\Downarrow$\\ 
                                                               $||D(a)-D(b)||_2$\\ 
                                                               $pair
                                                               \; label$ \\
                                                             \end{tabular}};
\draw [draw,<->] (cnn1) -- node {$w$} (cnn2);
\draw [draw,->] (patch1) -- node {$a$} (cnn1);
\draw [draw,->] (patch2) -- node {$b$} (cnn2);
\draw [draw,->] (cnn1) -- node [midway, above]  {$D(a)$} (loss);
\draw [draw,->] (cnn2) -- node [midway, below] {$D(b)$} (loss);
\end{tikzpicture}

\vspace{0.4cm}

% deep3
\begin{tikzpicture}[auto, node distance=2cm,>=latex']
  \node (patch1) at (0,0) {\includegraphics[width=.05\textwidth]{patch.png}};
  \node (patch2) [below = 0.45cm of patch1] {\includegraphics[width=.05\textwidth]{patch+.png}};
  \node (patch3) [below = 0.45cm of patch2]
  {\includegraphics[width=.05\textwidth]{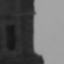}};

  \node (cnn1) [cnn, right = 0.5cm of patch1] {CNN};
  \node (cnn2) [cnn, below =0.5cm of cnn1] {CNN};
  \node (cnn3) [cnn, below =0.5cm of cnn2] {CNN};

  \node (loss) [loss, below right=0.25cm and 1.5cm of cnn1] {\begin{tabular}{c}
                                                               $Loss:
                                                               \;
                                                               function
                                                               \; of $\\
                                                               $\Downarrow$\\ 
                                                               $||D(p_1)-D(p_2)||_2$\\ 
                                                               $||D(p_1)-D(n)||_2$\\ 
                                                               $||D(p_2)-D(n)||_2$\\ 

                                                              \end{tabular}};
\draw [draw,<->] (cnn1) -- node {$w$} (cnn2);
\draw [draw,<->] (cnn2) -- node {$w$} (cnn3);
\draw [draw,->] (patch1) -- node {$p_1$} (cnn1);
\draw [draw,->] (patch2) -- node {$p_2$} (cnn2);
\draw [draw,->] (patch3) -- node {$n$} (cnn3);
\draw [draw,->] (cnn1) -- node [midway, above]  {$D(p_1)$} (loss);
\draw [draw,->] (cnn2) -- node [midway, below] {$D(p_2)$} (loss);
\draw [draw,->] (cnn3) -- node [midway, below] {$D(n)$} (loss);
\end{tikzpicture}

\caption{(top) Training of the siamese architecture. The loss is computed based on the
  distance in a positive or a negative patch pairs. (bottom) Training of the
  proposed {\bf PN-Net} architecture. The loss is based on all three
   distances within the triplet of patches. For details refer to
  Sec. \ref{sec:conj-deep-netw}.
} \label{fig:architectures}
\end{figure}

\subsection{Conjoined deep network loss functions}
\label{sec:conj-deep-netw}

In this section we first discuss the {\em Hinge Embedding} loss
\cite{1640964} commonly used in conjoined deep network architectures
and then we introduce our new loss function that is used in the optimization
of the proposed {\bf PN-Net} triplet based architecture.

\subsubsection{Hinge Embedding Loss}
The recent works related to deep learning of feature descriptors
utilize patch pairs and the {\em Hinge Embedding}.
\cite{1640964,simo2015deepdesc,ZagoruykoCVPR2015,DBLP:journals/corr/WangSLRWPCW14}. 
{\em Hinge Embedding} criterion was also used with triplets of data points to learn pose descriptors for 3D objects \cite{wohlhart15}.

Let $\mathcal{P}=\{p_L,p_R\}$ denote a patch pair and $\mathcal{L}\in\{-1,1\}$
a label indicating negative and positive pairs respectively. 
{\em Hinge Embedding} loss is then computed
\begin{equation}
  \label{eq:hinge-embedding-loss}
  l(\mathcal{P}) =
\left\{
	\begin{array}{ll}
		\Delta=||D(p_{L})-D(p_{R})||_2  & \mbox{if } \mathcal{L}=1 \\
		max(0,\mu-||D(p_{L})-D(p_{R})||_2)  & \mbox{if } \mathcal{L}=-1
	\end{array}
\right.
\end{equation}
Intuitively the hinge embedding loss penalizes positive pairs
that have large distance and negative
pairs that have small distance (less than $\mu$).

However as observed in \cite{simo2015deepdesc}, the majority of the
negative patch pairs ($\mathcal{L}=-1$) do not contribute to
the update of the gradients in the optimization process as their distance is already larger than $\mu$ parameter in
Eq. \eqref{eq:hinge-embedding-loss}. To address this issue hard negative mining was proposed to include more negative pairs in the training. The hardest negative training pairs were identified by their distance and a subset of these examples were re-fed to the network for gradient update in each iteration. 

\subsubsection{SoftPN loss}
We extend the idea of hard negative mining by incorporating both, positive and negative examples simultaneously in a new loss function.
 It uses a triplet of patches where two pairs represent a form of
 soft negative mining without the need for extra back propagation of specific hard negatives trough the network \cite{simo2015deepdesc}. 

 Any training triplet $\mathcal{T}=\{p_1,p_2,n\}$  includes two negative $\Delta(p_1,n) \;,\; \Delta(p_2,n)$ and one positive 
$\Delta(p_1,p_2)$ distance. Our proposed formulation of the loss function arises from the triplet CNN architecture 
illustrated  in Figure \ref{fig:architectures} (bottom). 
%Similarly to the second nearest neighbour distance ratio in \cite{Lowe:2004:DIF:993451.996342} 
It is based on the intuitive
idea that the smallest {\em  negative} distance within the triplet 
should be larger than the {\em positive} distance.  Ideally,
we require the positive distance to reach $0$ and the
two negative distances to increase towards $+\infty$. Therefore the {\em
  smaller} negative distance is the soft negative. Identification of such pair requires much less computation than mining of hard negatives and back propagating them through the network after every iteration. 
Formally our SoftPN objective is 
\begin{multline}
  \label{eq:deep3-objective}
   l(\mathcal{T}) = \large[ \big( \frac{e^{\Delta(p_1,p_2)}}{e^{min(\Delta(p_1,n),\Delta(p_2,n))}+e^{\Delta(p_1,p_2)}}\big)^{2}
   + \\
    \big(\frac{e^{min(\Delta(p_1,n),\Delta(p_2,n))}}{e^{min(\Delta(p_1,n),\Delta(p_2,n))}+e^{\Delta(p_1,p_2)}}  -1\big)^{2} \large]
\end{multline}

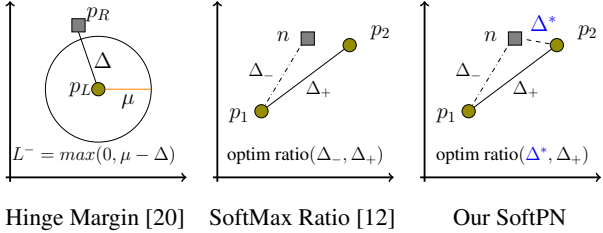
\begin{figure}
    \begin{subfigure}[b]{0.32\columnwidth}
        \centering
        \resizebox{\linewidth}{!}{
            \begin{tikzpicture}
              \draw (2,2) circle (1.2cm);
              \draw[-, very thin] (2,2) -- (1.55,3.3) node[midway,right] {\Large$\Delta$};
              \draw [draw=black, fill=olive] (2,2) circle (0.15cm) ;
              \node[draw=none] at (1.6,2) {\Large $p_L$};
              \draw [draw=black, fill=gray] (1.4,3.3) rectangle (1.7,3.6) ;
              \node[draw=none] at (2,3.7) {\Large $p_R$};
              \node[draw=none] at (1.9,0.5) {\large $ L^{-} = max(0,\mu - \Delta)$};
              \draw[-, very thin,orange] (2.15,2) -- (3.2,2)node[midway, below,black]{\Large $ \mu$};
              \draw[->, very thick] (-.1,0) -- (4,0) node[below] {$\text{Α}$};
              \draw[->, very thick] (0,-.1) -- (0,4) node[left]{$\text{Β}$};
            \end{tikzpicture}
        }
        \caption*{Hinge Margin \cite{simo2015deepdesc}}
            \end{subfigure}
    \begin{subfigure}[b]{0.32\columnwidth}
    \centering
        \resizebox{\linewidth}{!}{
            \begin{tikzpicture}

              % \draw[-, very thin] (2,2) -- (1.55,3.3) node[midway,left] {\Large$\Delta$};

              \draw[-, very thin] (1,1.5) -- (3,3);
              \draw[dashdotted, very thin] (1,1.5) -- (1.9,3);
              \draw [draw=black, fill=olive] (1,1.5) circle (0.15cm) ;
              \node[draw=none] at (0.5,1.4) {\Large $p_1$};
              \draw [draw=black, fill=olive] (3,3) circle (0.15cm) ;
              \node[draw=none] at (3.7,3.3) {\Large $p_2$};
              \draw [draw=black, fill=gray] (1.9,3) rectangle (2.2,3.3) ;
              \node[draw=none] at (1.5,3.2) {\Large $n$};
              \node[draw=none] at (2.3,2) {\large $\Delta_{+}$};
              \node[draw=none] at (1,2.4) {\large $\Delta_{-}$};
              \node[draw=none] at (2,0.5) {\large $\text{optim ratio}(\Delta_{-},\Delta_{+})$};
              \draw[->, very thick] (-.1,0) -- (4,0) node[below] {$\text{Α}$};
              \draw[->, very thick] (0,-.1) -- (0,4) node[left]{$\text{Β}$};
            \end{tikzpicture}
        }
        \caption*{{\small SoftMax Ratio \cite{DBLP:journals/corr/HofferA14}}}   
    \end{subfigure}
    \begin{subfigure}[b]{0.32\columnwidth}
        \centering
        \resizebox{\linewidth}{!}{
            \begin{tikzpicture}
  \draw[-, very thin] (1,1.5) -- (3,3);
              \draw[dashdotted, very thin] (1,1.5) -- (1.9,3);
              \draw[dashed, very thin] (1.9,3.2) -- (3,3);
              \draw [draw=black, fill=olive] (1,1.5) circle (0.15cm) ;
              \node[draw=none] at (0.5,1.4) {\Large $p_1$};
              \draw [draw=black, fill=olive] (3,3) circle (0.15cm) ;
              \node[draw=none] at (3.7,3.3) {\Large $p_2$};
              \draw [draw=black, fill=gray] (1.9,3) rectangle (2.2,3.3) ;
              \node[draw=none] at (1.5,3.2) {\Large $n$};
              \node[draw=none] at (2.3,2) {\large $\Delta_{+}$};
              \node[draw=none] at (1,2.4) {\large $\Delta_{-}$};
              \node[draw=none,blue] at (2.7,3.5) {\Large $\Delta^{*}$};
              \node[draw=none] at (2,0.5) {\large $\text{optim
                  ratio}({\color{blue} \Delta^{*}},\Delta_{+})$};
              \draw[->, very thick] (-.1,0) -- (4,0) node[below] {$\text{Α}$};
              \draw[->, very thick] (0,-.1) -- (0,4) node[left]{$\text{Β}$};
            \end{tikzpicture}
        }
        \caption*{Our SoftPN}
    \end{subfigure}
\caption{Illustration of loss functions. Note that the proposed SoftPN
  loss (right) evaluates the smallest negative distance $\Delta^{*}$ within a triplet.} 
\label{fig:objectives}
\end{figure}

Given $\Delta^{*}=min(\Delta(p_1,n),\Delta(p_2,n))$ as the soft negative distance, the goal of the loss function is to force $\Delta^{*}$
to be  larger than $\Delta(p_1,p_2)$. Note that unlike in the {\em
  Hinge Embedding} loss, negative distances always contribute to the
 optimization and in contrast to the previous works on triplet based learning
\cite{DBLP:journals/corr/HofferA14,DBLP:journals/corr/WangSLRWPCW14},
our negative loss includes both negative distances
$\Delta(p_1,n),\Delta(p_2,n)$. 
%By using the $min(D_{qq^{-}},D_{q^{+}q{-}})$ we enforce a kind of weak {\em hard negative mining} inside each triplet. 
%Intuitively, the irrelevant sample $n$ in the triplet is extremely unlikely to be lying in the center of an $D-sphere$ in the $D-dimensional$ feature space, equally distanced from the two positive samples $p_1$ and $p_2$. 
The {\em SoftMax Ratio} objective introduced in \cite{DBLP:journals/corr/HofferA14} is 
based on triplets and ratios, but does not include the evaluation of
the $\Delta^{*}$ distance inside the triplet. 
An illustration of the the hinge loss objective
\cite{ZagoruykoCVPR2015,simo2015deepdesc,DBLP:journals/corr/WangSLRWPCW14}
, the {\em SoftMax Ratio} \cite{DBLP:journals/corr/HofferA14}
and the proposed SoftPN loss function is presented in Figure \ref{fig:objectives}.

\begin{figure}
    \centering
    \begin{subfigure}[b]{0.24\columnwidth}
        \centering
        \includegraphics[width=\columnwidth]{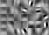}
    \end{subfigure}%
    \hspace{0.2cm}
    \begin{subfigure}[b]{0.64\columnwidth}
        \centering
        \includegraphics[width=\columnwidth]{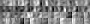}
    \end{subfigure}
    \vspace{0.05cm}

    \hspace{-0.2cm}
    \begin{subfigure}[b]{0.24\columnwidth}
        \centering
        \raisebox{0.6ex}{\includegraphics[width=\columnwidth]{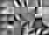}}
    \end{subfigure}%
    \hspace{0.2cm}
    \begin{subfigure}[b]{0.635\columnwidth}
        \centering
        \includegraphics[width=\columnwidth]{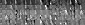}
    \end{subfigure}
    \caption{ The weights learned in the first layer (left)  and in
      the second layer (right)  of the
      CNN. (Top) is our PN-Net and (bottom) is DeepCompare.}
\label{fig:deep_vs_sergey}
\end{figure}

\begin{figure}
    \centering
    \begin{subfigure}[b]{0.5\columnwidth}
        \centering
        \includegraphics[trim=15 1 15 1,width=\columnwidth]{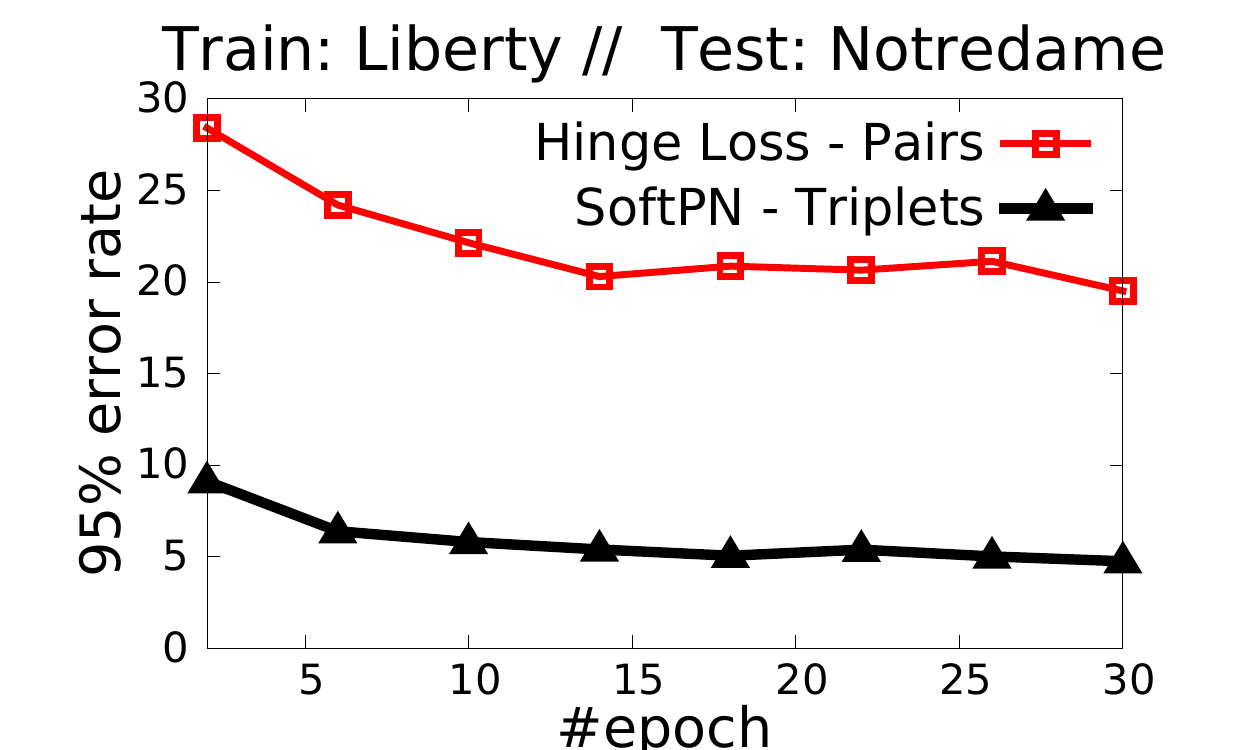}
        % \caption{Lorem ipsum}
    \end{subfigure}%
    \begin{subfigure}[b]{0.5\columnwidth}
        \centering
        \includegraphics[trim=15 1 15 1,width=\columnwidth]{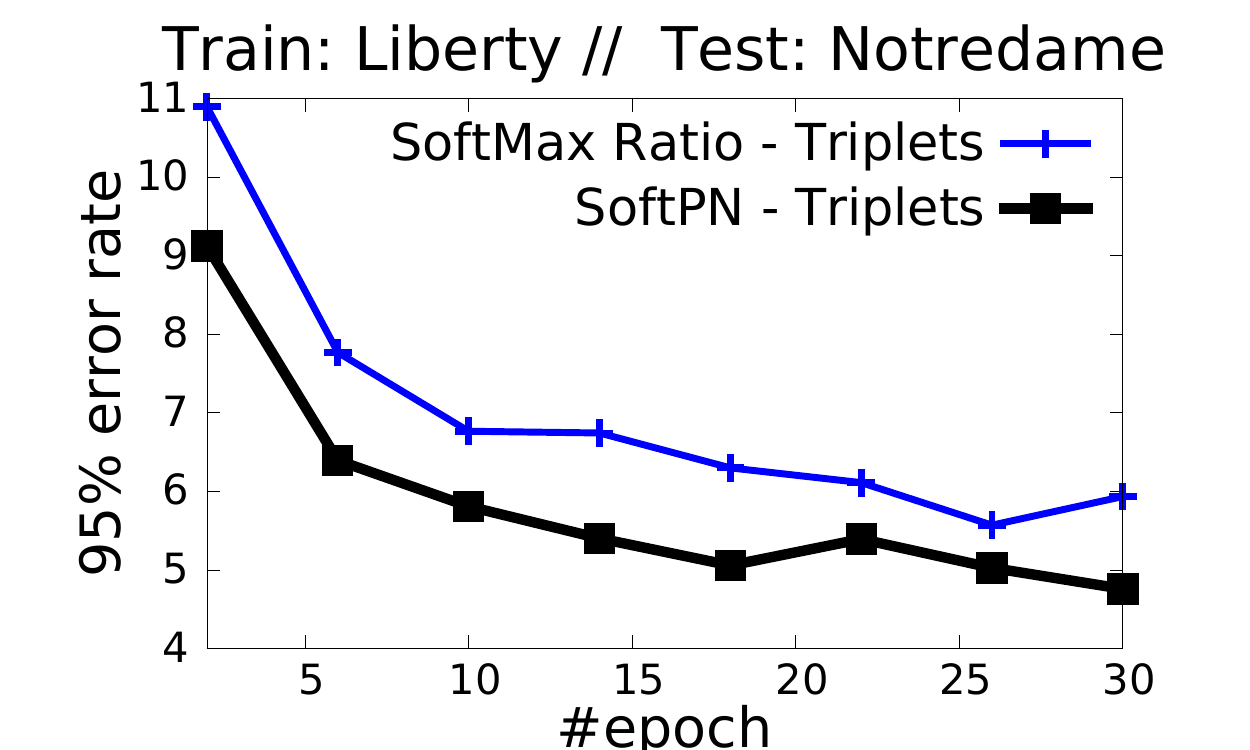}
    \end{subfigure}
    \caption{(left) The proposed SoftPN loss compared with the Hinge
      Margin Loss used commonly in learning deep descriptors, when
      using the same underlying CNN (Table
      \ref{tab:cnn-architecture}), and trained on exactly the same
      data. Note the significant improvements of performance.(right)
      The effect of the local negative mining inside each triplet. We
      see that the use of the soft negative $\Delta{*}$ distance
    inside each triplet leads to lower error rates.}
\label{fig:deep_vs_siamese_vs_tripletnet}
\end{figure}

Figure \ref{fig:deep_vs_sergey} visualises two layers of the CNN for the proposed PN-Net as well as for DeepCompare~\cite{ZagoruykoCVPR2015}. Convolutional filters of PN-Net seem to be more smooth e.g. more regularised compared to DeepCompare. We believe it is the effect of simultaneous use of positive and negative pairs in the loss function during training.
In Figure \ref{fig:deep_vs_siamese_vs_tripletnet} we compare the effect different loss functions discussed in this section have on the matching performance of learnt descriptors. 
We plot the 95\% error in matching patches from patch dataset 
\cite{BHW10},   and show how it decreases with each training epoch. A first observation
is that the triplet based learning
is significantly better for learning feature descriptors than the state of the art
siamese pair based learning. 
This is demonstrated by the large difference between the siamese \& hinge
margin architecture compared to the proposed PN-Net architecture. The final error rate is 15\% lower for SofPN. Next, by comparing results for SoftMax and SoftPN to hinge loss we conclude that using triplets of data points as training examples leads to much better results than using pairs. 

We
also note that the proposed SoftPN loss function outperforms SoftMax Ratio function, due to the soft negative mining by
$\Delta^{*}$ distance.  
Moreover,  already the first epoch (\ie after $2 mins$) of training a local feature descriptor with the proposed
method  leads to  the matching error rate of $9\%$ when training the network with
the Liberty dataset and testing in the Notredame dataset. This is much lower error than many recent descriptors achieve after extensive training as we show in section \ref{s:results}. 

%We can also put this in the introduction? It's like DSP-SIFT intro with the overexposed percentage differences. 

Another important note is that for training and comparing different loss functions we use the same set of patches.
There are three times more pairs in this set than triplets therefore the siamese network with hinge loss uses three times more training examples than TripletNet with SoftMax and PN-Net with SoftPN. This is a significant advantage for the siamese network, yet the results indicate that the training is much more effective with triplets. It is also more efficient due to use of less training examples.

\subsection{Implementation details}

PN-Net and the siamese networks used the same
underlying CNN  in the experiments
above. Also, unlike in
\cite{ZagoruykoCVPR2015} and \cite{Han_2015_CVPR} there is no extra
layer that learns a distance metric between the two patches. Our simplified CNN architecture allows more efficient descriptor extraction and the use of $L_2$ norm for matching. 
Thus, a  descriptor is obtained by processing a patch by a single branch of the network.
The parameters of the network used in all our experiments are presented in Table
\ref{tab:cnn-architecture}.

\begin{table}[h]
  \begin{center}
    \caption{Architecture of the CNN used in the experiments. Patches
      of size $32\times 32$ are used as input. The number in each
      convolutional layer denotes the number of the output planes that
    the convolution produces.}
    \label{tab:cnn-architecture}
    \begin{tabular}{cc}
      \toprule
      Layer \# & Description\\
      \midrule
      1 & Spatial Convolution(7,7) $\rightarrow$ 32 \\
      2 & Tanh \\
      2 & MaxPooling(2,2) \\
      3 & Spatial Convolution(6,6) $\rightarrow$ 64 \\
      4 & Tanh \\
      5 & Linear $\rightarrow$ $\{128,256\}$ \\
      6 & Tanh \\
      \bottomrule
    \end{tabular}
  \end{center}
\end{table}

Our implementation is in \texttt{Torch} \cite{collobert:2011c}. The
training is done using $\approx 1.2M$ triplets generated on-the-fly
using the patches from \cite{BHW10}. In contrast to how CNNs are typically trained including DeepCompare and MatchNet we do not use data augmentation. This is to make the training more efficient and to demonstrate that the improvements result from the approach and not from larger number of patches used for training.

In forming the triplets we choose
randomly a pair of patches from the same 3D point, and subsequently we
complete the triplet with a randomly chosen patch from another 3D
point. This is in contrast to other works where carefully designed
schemes of choosing the training data are used in order to enhance the
performance~\cite{DBLP:journals/corr/WangSLRWPCW14,Han_2015_CVPR}.

For the optimization the
Stochastic Gradient Descend \cite{bottou-tricks-2012} is used, and the
training is done in batches of $128$ items, with a learning
rate of $0.1$, momentum of $0.9$ and weight decay of $10^{-6}$.

The convolution
methods are  from the NVIDIA cuDNN library
\cite{DBLP:journals/corr/ChetlurWVCTCS14}. 
The training of a single epoch with $\approx 1.2M$ training triplets
takes approximately $2$ minutes in an NVIDIA Titan X GPU.

It is worth noting that the CNN used in our experiments consists of
only two layers, while all of the other state-of-the art deep feature
descriptors consist of 3 layers and above
\cite{ZagoruykoCVPR2015,simo2015deepdesc,Han_2015_CVPR}. Several other
implementation variants could be added such as using different
non-linearity layers (e.g. ReLU as in
\cite{Han_2015_CVPR,ZagoruykoCVPR2015}), extra normalization layers,
but the main focus of our work is to show the effect of learning local
features with triplets coupled with the SoftPN loss function. Sample results
for other network configurations are presented in the supplementary material.

%%% Local Variables:
%%% mode:latex
%%% TeX-master: "egpaper_for_review"  ***
%%% End:

\section{Experimental evaluations}
\label{s:results}
In this section we evaluate the proposed local feature descriptor within the two most popular benchmarks in the field of local descriptor matching. We compare our method to SIFT \cite{Lowe:2004:DIF:993451.996342}, Convex optimization \cite{simonyan2014learning} the recently introduced MatchNet \cite{Han_2015_CVPR}
and DeepCompare \cite{ZagoruykoCVPR2015} descriptors which are
currently the state of the art in terms of matching accuracy.  
The original code available from the authors was used in all the experiments. Code for other  recent methods e.g. DSP-SIFT was not available at the time of this experiment.
 
Note that for a fair comparison, we use the siamese architectures similar to the one in
DeepCompare, but we do  not use the multi-scale {\em 2ch architectures}. Multi-scale approaches uses multiple patches from each example, with one
being a cropped sub-patch around the center. This introduces
information from different samples in the size-space and it has been shown to lead to significant improvements in terms of
matching accuracy \cite{DBLP:journals/corr/DongS14}. Such approach can be uses for various descriptors (e.g. MatchNet-2ch, PN-Net-2ch etc.).

It would be interesting to evaluate the effect of mining in terms of a
siamese network learning as it was proposed in
\cite{simo2015deepdesc}, however the implementation is not available
yet.

\begin{table*}[ht]
\caption {Results form the Photo-Tour dataset \cite{BHW10}. Numbers
  are reported in terms of {\em error at 95\% correct}. {\bf Bold}
  numbers indicate the best performing descriptor. Note the
  significant reduction in dimensionality by {\bf PN-Net}, together
  with the improvements over the state-of-the art results.} 
\label{tab:benchmark_brown} 
\begin{tabular}{ccccccccc}
\toprule 
    Training& & Notredame & Liberty & Notredame  & Yosemite & Yosemite
  & Liberty  & \\
    % \midrule
    Testing&  & \multicolumn{2}{c}{Yosemite} & \multicolumn{2}{c}{Liberty} & \multicolumn{2}{c}{Notredame} \\
    \midrule
    Descriptor & \# features & & & & & & &mean  \\
    \midrule
    SIFT \cite{Lowe:2004:DIF:993451.996342} & 128 & \multicolumn{2}{c}{27.29} & \multicolumn{2}{c}{29.84}
                                    & \multicolumn{2}{c}{22.53} &
                                                                  26.55 \\
    ConvexOpt \cite{simonyan2014learning} & $\approx80$ & 10.08 & 11.63
                                    & 11.42 & 14.58 & 7.22 & 6.17 &
                                                                    10.28 \\
    DeepCompare {\em siam} \cite{ZagoruykoCVPR2015}& 256 &
                                                                      13.21
                          & 14.89 & 8.77 & 13.48 & 8.38 & 6.01 & 10.07
  \\
     \hspace{1cm} {\em pseudo-siam} \cite{ZagoruykoCVPR2015}& 256 &
                                                                      12.64
                          & 12.5 & 12.87 & 10.35 & 5.44 & 3.93 & 9.62\\
    MatchNet \cite{Han_2015_CVPR} & 512 & 11 & 13.58
                                    & 8.84 & 13.02 & 7.7 & 4.75 & 9.82\\
    \hspace{1cm}  {\em no bottleneck} \cite{Han_2015_CVPR} & 4096 & 8.39 & 10.88
                                    & {\bf 6.90} & 10.77 & 5.76 & 3.87 & 7.75\\
  {\bf PN-Net} & {\bf 128} & {\bf 7.74}& {\bf 9.55}
                                    & 8.27  & {\bf 9.76} & {\bf 4.45}
  & {\bf 3.81} & {\bf 7.26}\\
 {\bf PN-Net} & {\bf 256} & {\bf 7.21 }& {\bf 8.99}
                                    & 8.13  & {\bf 9.65 } & {\bf 4.23 }
  & {\bf 3.71} & {\bf 6.98}\\
    \bottomrule
\end{tabular}
\end{table*}

\subsection{Photo Tour dataset}
We first evaluate the performance in terms of matching accuracy in
distinguishing positive from negative patch pairs on the Photo Tour
dataset \cite{BHW10}. This dataset consists of three subsets {\em
  Liberty},{\em Yosemite} \& {\em Notredame} containing more than 500k patch pairs extracted around specific feature points.
We follow the protocol proposed in  \cite{BHW10}
where the ROC curve is generated by thesholding the distance scores
between patch pairs. The number reported here is the false positive
rate at 95\% true positive rate. For the evaluation we use the $100K$
patch pairs proposed by the authors.
%The list of distances together with
%the matching label of the patch (positive or negative) are used in
%order to compute a ROC curve. Intuitively this benchmark measure how
%good a descriptor is in separating positive from negative patch
%pairs. We report our findings in terms of error rate at 95\% true
%positive matches.

The results for each of the combinations of training and testing using
the three subsets of the Photo Tour dataset are shown in
Table~\ref{tab:benchmark_brown}. The average
across all possible combinations is also shown.
Our PN-Net outperforms the state of the art for a single scale siamese CNN
architecture (MatchNet). Moreover, our network does not learn an explicit distance metric and the final layer gives 128 dimensional descriptor in contrast to 4096 of MatchNet. Note that the 
performance gain is even greater when comparing to DeepCompare descriptors from
\cite{ZagoruykoCVPR2015}, event though their dimensionality is still twice larger than from PN-Net. Moreover, our descriptor
outperforms all the others, except in the combination of training in
Notredame and testing in Liberty. 
%A possible explanation for this
%phenomenon is the fact that the Notredame patches are very high in
%detail, something that might discourage the SoftPN criterion. 
We can also observe that there is not much difference in 
performance when comparing the $128$ with the $256$ variants of
PN-Net. It is important to mention that the proposed descriptor achieves state of
the art performance without any data augmentation during training,
 in contrast to the competing CNN based methods \cite{ZagoruykoCVPR2015,Han_2015_CVPR}. 
Smaller network and no data augmentation or multi-scale examples leads to much faster learning, yet  our PN-Net and SoftPN loss achieves   similar or better performance after 1 epoch ($2min$) than DeepCompare \cite{ZagoruykoCVPR2015}   after 2 days of training. 
Full plots of the ROC curves can be found in the supplementary
material.

\subsection{Oxford image sequences}

We also test the performance of the proposed descriptor in terms of
matching local features between two images based on the benchmark from
\cite{schmid2003performance}.

\noindent{\bf Evaluation protocol.} As discussed in the introduction the original protocol has been loosely followed in various papers leading to not comparable results. 
%We are interested in standardizing the process of benchmarking based on this dataset, since currently a great number of different detectors and evaluation methods are used. 
To address this issue, we draw ideas from the successful Photo Tour benchmark. 
To increase the size of the dataset we complement the eight sequences with additional seven that include illumination, rotation and scale changes. The images are acquired in a similar way to the original eight with pair-wise image homographies for establishing correspondence ground truth.
Note that unlike \cite{FDB14}, the sequences are not generated artificially by
warping single images to various transformations, but
they come from naturally captured images together with the noise introduced by varying imaging conditions. 
Next, we apply an interest point detector to identify regions with varying image content.  
In contrast to the Photo Tour~\cite{BHW10} that extracted keypoints with scale-invariant DoG, we use affine invariant Harris-Affine. This makes the patches complementary to the Photo Tour and introduces  a different type of noise resulting from the affine detector inaccuracy.
We establish correspondence ground truth using the homographies and the  overlap error from~\cite{schmid2003performance}.
We consider two points in correspondence if the overlap error between the detected regions is less than 50\%. Note that a region from one image can be in correspondence with several regions from the other image. 
By reducing each image pair to a collection of patches, the benchmark will eliminate
some of the varying factors such as the choice of the interest point detector, the number  of extracted patches, their sizes, etc.
The data includes 15 sequences, each consisting of 6 images with increasing degree of change in viewing conditions.
Each image has an associated set of ~1k patches. A pair of images has an associated text file indicating the overlap error between the patches that overlap by at least 50\%. 
This data will be available online and together with Photo Tour will allow to test additional properties of new descriptors such as robustness to different type of noise and generalisation from one dataset to the other.
Note that the results for Oxford data presented here  are not directly comparable to those reported in other papers. However this has been the case for many papers and we believe that using patches will allow converging to repeatable experiments on this data.
The descriptors performance is evaluated in terms of nearest neighbour matching and the results are presented with 
 precision-recall curves as it was originally proposed in
\cite{schmid2003performance}. 
 More specifically, for each patch from the left image we find its nearest neighbour in
the right image. Based on the ground truth overlap we can distinguish between false positives and true
positives, and generate  precision-recall curves. Detailed
description of the dataset and the benchmarking protocol can be found in the
supplementary material. 

\begin{figure}[t]
    \centering
        \hspace{-0.1cm}\includegraphics[trim=3cm 0 3cm 0,width=0.5\columnwidth]{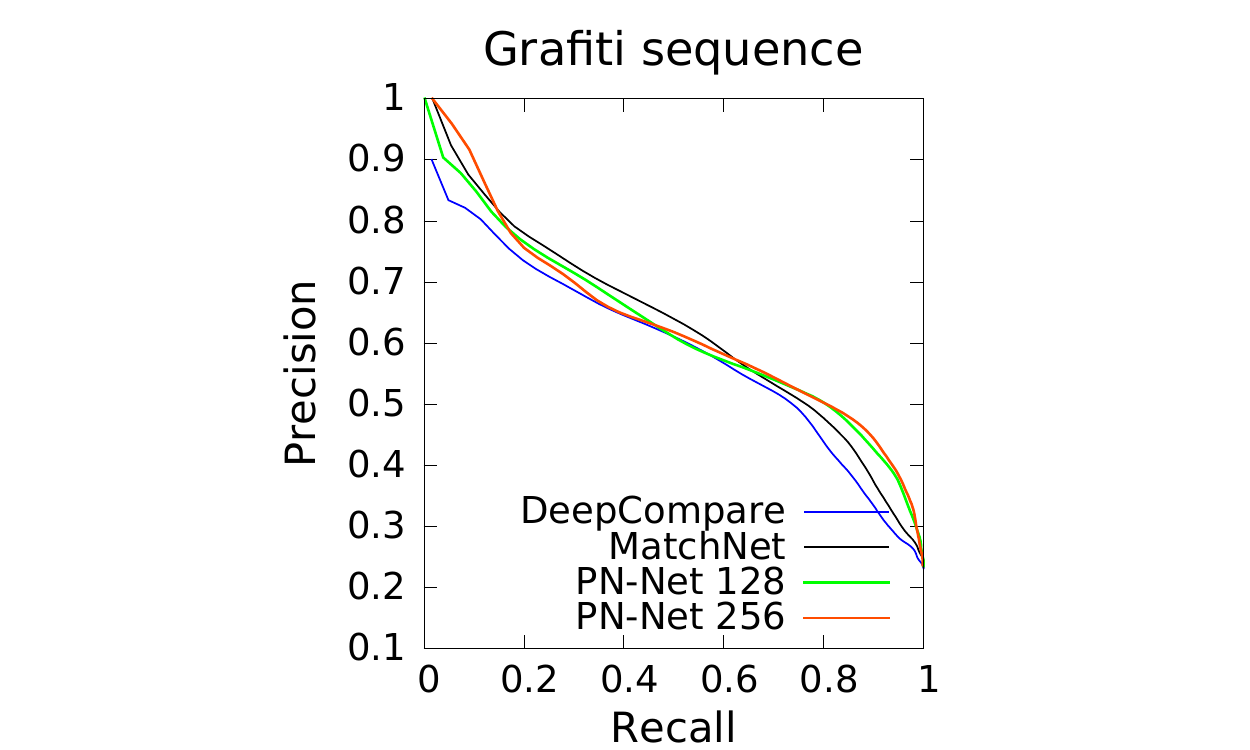}
        \includegraphics[trim=3cm 0 3cm 0,width=0.5\columnwidth]{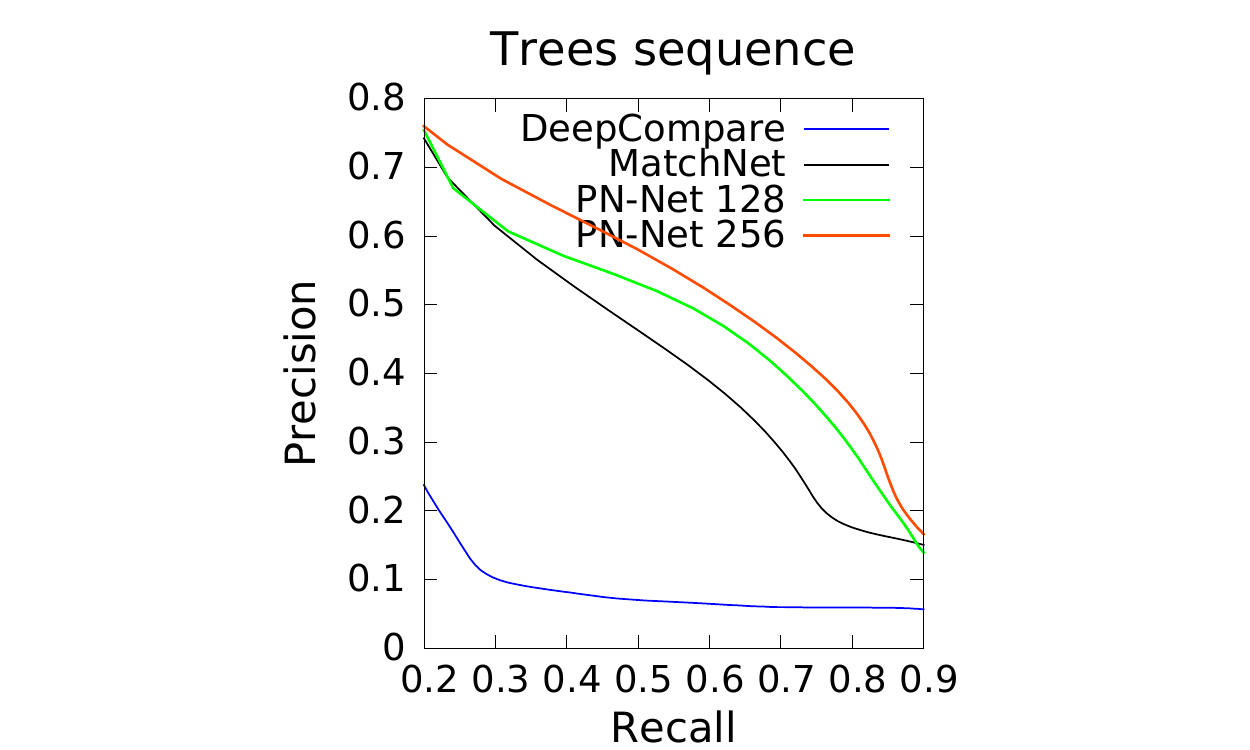}
    \caption{ Precision-Recall curves for matching performance tested on Oxford data.  Results for the two challenging sequences: $graffiti$ (left) with affine deformations  and $trees$ (right) with blurr and out of focus images.    Our PN-Net gives top scores for both sequences.}
\label{fig:grafpr}
\end{figure}

In  this experiment all the CNN networks are trained using Libery data from Photo Tour. In Figure \ref{fig:grafpr} we present the precision-recall curves across
all the pairs of  $graffiti$  and $trees$ images, which are considered  the most 
challenging sequence in the dataset. MatchNet and the
proposed PN-Net are  close in terms of the area under the curve,
although PN-Net outperforms MatchNet and DeepCompare in both sequences in particular in $trees$. Clearly DeepCompare struggles with image blurr which confirms the observations in \cite{ZagoruykoCVPR2015}. 
 We include precision recall curves for the other sequences in the suplementary material.

%, together with their ground truth
%correspondences in terms of feature overlap. For any two images $L,R$, let
%$N_L$ be the number of feature points extracted from the left image,
%and $N_R$ from the right. We introduce as ground truth the overlap
%matrix, which is an $N_l \times N_R$ overlap matrix that contains the
%overlaps between all the combinations of feature points from the two
%images. We augment the original dataset from
%\cite{schmid2003performance} with more sequences, and we release a
%benchmark containing extracted patches and ground truths. 

\begin{figure}[t]
    \centering
    \begin{subfigure}[b]{1.0\columnwidth}
        \centering
        \includegraphics[width=\columnwidth]{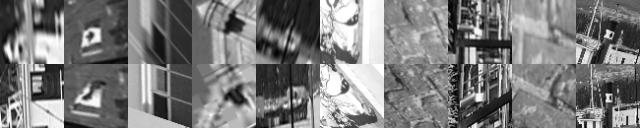}\\
        \vspace{3pt}
%        % \caption{Lorem ipsum}
%    \end{subfigure}%
%    \begin{subfigure}[b]{1.0\columnwidth}
%        \centering
        \includegraphics[width=\columnwidth]{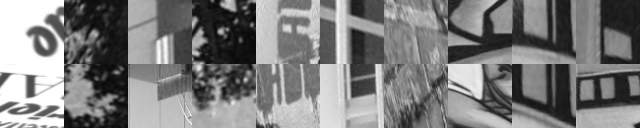}
        % \caption{Lorem ipsum, lorem ipsum,Lorem ipsum, lorem ipsum,Lorem ipsum}
    \end{subfigure}
    \caption{ Examples of true (top) and false positive (bottom)
      nearest neighbour matching in a large patch dataset. Note the
      extreme variations that the proposed descriptor can cope with.}
\label{fig:true_and_false_positives}
\end{figure}

Some examples of true and false positive matches between several image pairs are shown in Figure \ref{fig:true_and_false_positives}. True positives are
shown in the top row, and false positives in the bottom. The patches show extreme affine deformation, scale changes, rotation error due to inaccuracy of the Harris-Affine detector as well as the interpolation noise resulting from normalisation from ellipses to circles.   Some of the correct matches by PN-Net are impressive given that the network was not trained on such errors as they don't occur in Photo Tour data. Also some of the false positives exhibit some visual similarity to be considered as reasonably explicable.

\begin{figure*}
\centering
\vspace{-3.5cm}
\includegraphics[width=\textwidth]{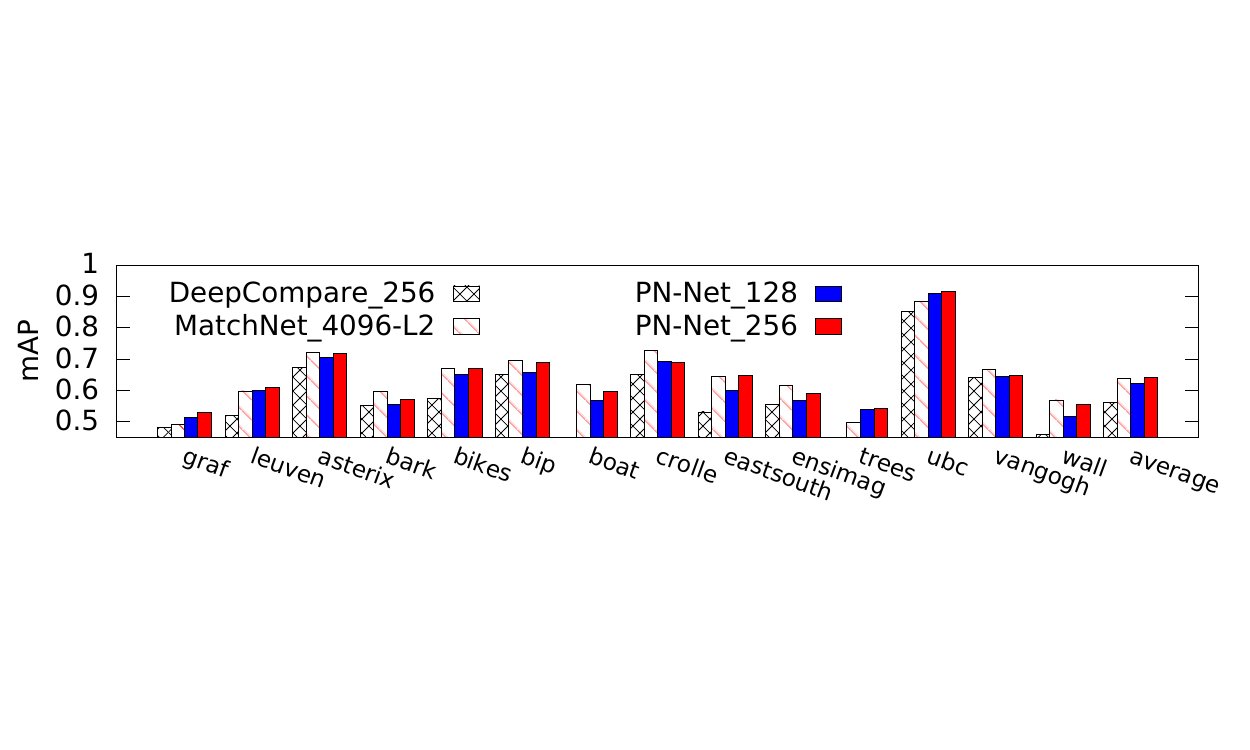}
\vspace{-4cm}
\caption{Mean average precision (mAP) in terms of nearest neighbor descriptor
  matching based on the protocol from \cite{schmid2003performance}.  }
\label{fig:oxford}
\end{figure*}

To compress the results we use the mean average precision (mAP) which is defined as
the area under the precision recall curve. The mAP for different methods and image sequences is presented in Fig. \ref{fig:oxford}.
First observation is that the scores for the new images fall  between the $graf$ and the $ubc$ sequence, which are typically the most and the least challenging ones in the original data.
More importantly, the  PN-Net\_256 outperforms the matching
results of MatchNet by $2\%$ in terms of mAP averaged across all the
sequences, even though MatchNet descriptor consists of 4095 dimensions.
In particular, in sequences with affine deformations and blurr ($graf$ and $trees$). 
Interestingly, the results for scale changes ($asterix$, $bark$, $croll$ and $boat$) are inverse.
This clearly shows that multi-scale training has an impact on the results and further improvements can be made if PN-Net is trained with data augmentation. Note that for this NN matching experiment with MatchNet we used the  L2 distance and the feature vector resulting from an intermediate layer before the final metric layer. This is because the metric layer of
MatchNet did not produce meaningful results due to the fact that it is
mainly trained to act as a classifier rather than an accurate distance measure.
It is worth noting the significant improvement
that PN-Net gives  compared to the descriptor of similar
dimensionality i.e. DeepCompare \cite{ZagoruykoCVPR2015}, which clearly falls behind the other two in this experiment.
Another positive point is that the generalisation from Liberty to Oxford is good for PN-Net and MatchNet, which may indicate that there is no over-fitting to a specific type of images or interest point detectors.

%We previously exhibited that in the Photo Tour benchmark, which consists of pairs of positive and negative patches and th  benchmarking protocol aims to discover how well separated they are in the descriptor feature space, there is no significant difference between the $128$ and $256$ variants of the proposed NP-Net. However, we from Figure~\ref{fig:oxford} that a significant gain in performance can be gained from doubling the final descriptor dimensions. This can be explained due to the fact that the Oxford dataset matching protocol is concerned with nearest neighbour matching between two images, thus it is crucial for the descriptor to be able to handle this case. 

%This further supports the observation that CNN generalise well across different datasets.

\subsection{Computational efficiency}

One of the main motivations behind this work, was the need for a fast
and practical feature descriptor based on CNN. The proposed PN-Net
descriptor is very efficient  in  terms of both training and extraction time. 

\begin{figure}[!ht]
  \begin{subfigure}[c]{.6\columnwidth}
    \centering
        \raisebox{-1.1\height}{\includegraphics[trim=15 1 15 1,width=\columnwidth]{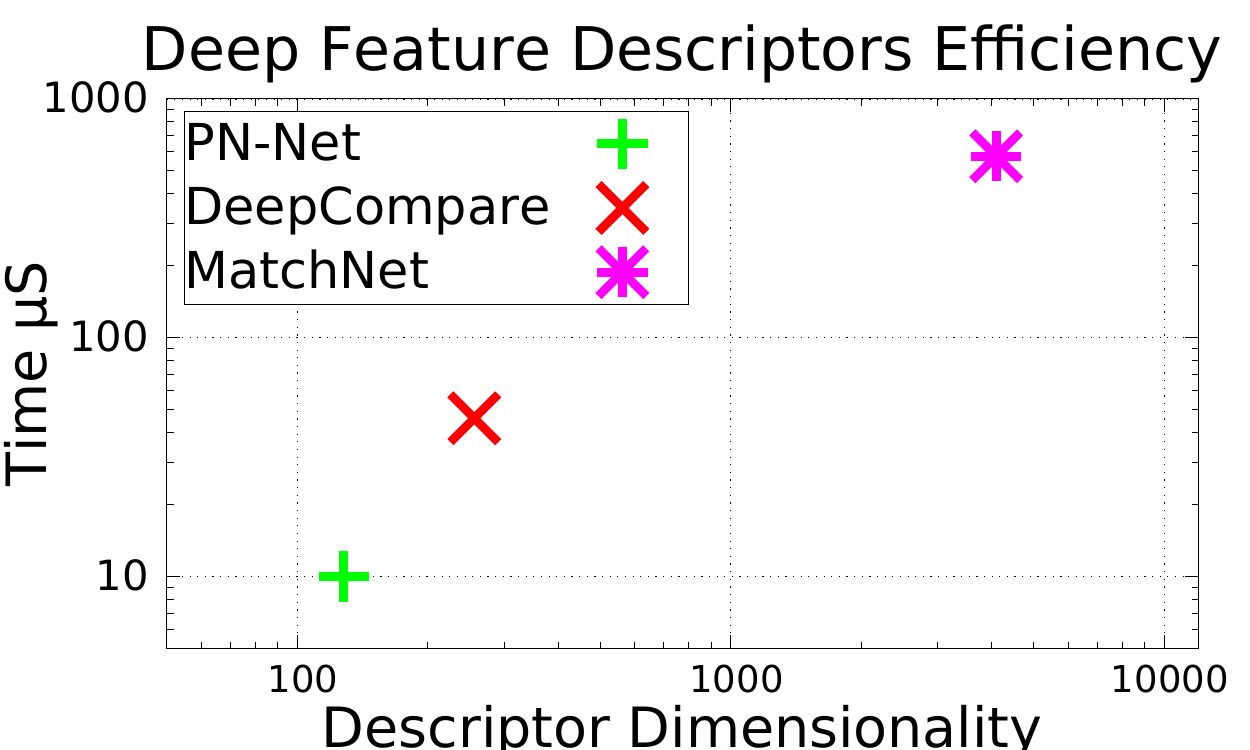}}
  \end{subfigure}\hfill
  \begin{subtable}{.38\columnwidth}
    \centering
    \begin{tabular}[C]{|l|c|} 
 \multicolumn{2}{ c }{$\approx \text{time } \mu S$}\\
      % &$\approx time$ per descriptor $\mu S$\\
      \hline
      \small SIFT (CPU) & 400 \\  
      \small BRIEF (CPU) & 3 \\  
      \small { PN-Net} & { 10} \\
     \small MatchNet  & 575 \\  
      \small DeepCompare  & 46 \\  
      \hline
    \end{tabular}
  \end{subtable}
  \caption{(left) Dimensionality and efficiency  of the proposed PN-Net feature compared to other methods. We report  the time required to extract a descriptor from a single input patch in $\mu S$. Note
    that both axes are in logarithmic scale. (right) We note that the
    GPU version of PN-Net approaches the efficiency of BRIEF which is
    the fastest CPU descriptor available.}
  \label{fig:efficiency}
\end{figure}

In Figure \ref{fig:efficiency} (left) we present results on the extraction
times w.r.t. dimensionality.  For the GPU implementations of the deep networks all
experiments were done with an NVIDIA GTX TITAN X GPU. 
We can see that when compared with the recently proposed deep feature
descriptors, the proposed PN-Net is both faster and smaller in
dimensionality, while at the same time performs better in the
benchmarks.

From the results in Figure \ref{fig:efficiency} (right) we can
conclude that the GPU version of the proposed PN-Net descriptor is
close to the speed of the  CPU implementation of BRIEF. This
gives a significant advantage over the previously proposed descriptors
 and makes CNN based descriptors applicable to practical problems with large datasets. 

Note that several works have attempted to port SIFT to  GPU \cite{sihna2006gpu}, with
speedups ranging from 5 to 20 compared to the CPU version. Even when considering such speedups, the proposed descriptor is still faster to compute mainly due to the convolutional operations libraries
\cite{DBLP:journals/corr/ChetlurWVCTCS14}.

Note that the proposed PN-Net descriptor also has an advantage in
the computational efficiency during training time. While other works
mention that their optimizations take from several hours
\cite{simonyan2014learning} to 2 days
\cite{ZagoruykoCVPR2015,Han_2015_CVPR}, our work reaches state of the
art performance in 100-200 training epochs, which translates to 2-5
hours training on a single GPU. Surprisingly even after a single
epoch, \ie after two minutes of training, we get a descriptor very
close to the state of the art. 

%%% Local Variables:
%%% mode:latex
%%% TeX-master: "egpaper_for_review"  ***
%%% End:

\section{Conclusion}
This work introduced a new approach to training CNN architecture for extracting local image descriptors in the context of nearest neighbour matching.  It is based on the recent advancements in the area of convolutional neural networks and deep
learning.  It makes use of the ideas introduced in the field of distance metric learning and online boosting by training with 
positive and negative constraints simultaneously. 

We have introduced a novel loss function SoftPN that is based on triplets of
patches extracted from feature points. It incorporates the idea of hard negative mining within the loss function thus avoid   mining  and retraining of the network after each iteration.  The experimental results show that SoftPN leads to faster convergence and lower error than hinge loss or SoftMax ratio.

Moreover, the results show that using triplets for training results in a better descriptor and faster learning.
The networks can be made simpler, trained with less examples and extract descriptors with a speed comparable to BRIEF. Also the dimensionality can be significantly reduced compared to other CNN based descriptors.
We believe that due to these properties the proposed network is less prone to over-fitting. 
This is supported by the results showing good generalisation properties.

In addition, we proposed a new  protocol for evaluating feature matching that aims to correct the significant discrepancies between the protocols used in various experiments and ambiguities in interpreting the results of descriptor performance evaluations in future papers.   We added more image sequences to the frequently used Oxford dataset \cite{schmid2003performance}.
The dataset, the ground truth and the source code of the proposed PN-Net
descriptor are available from \url{https://github.com/vbalnt/pnnet}.

We believe that the proposed PN-Net descriptor with its
very efficient computation, low memory requirements  and state of the art performance will
enable new real-time applications that are based on a new family of
highly accurate but extremely fast deep feature descriptors. We also believe that this work sets a positive example that there is no need to compromise the efficiency of the descriptor to achieve state of the art performance with CNN architectures.

{\small
\bibliographystyle{ieee}

}

\end{document}